\documentclass[10pt, a4paper]{article}
\usepackage{lrec}
\usepackage{multibib}
\newcites{languageresource}{Language Resources}
\usepackage{graphicx}
\usepackage{tabularx}
\usepackage{soul}
\usepackage[dvipsnames]{xcolor}

\usepackage{arydshln}

\usepackage{flushend}
\usepackage[ruled,vlined]{algorithm2e}

\usepackage{epstopdf}
\usepackage[utf8]{inputenc}

\usepackage{hyperref}
\usepackage{xstring}

\title{Automatic Spanish Translation of the SQuAD Dataset for Multilingual Question Answering\\ }

\name{Casimiro Pio Carrino, Marta R. Costa-jussà, José A. R. Fonollosa}

\address{TALP Research Center\\
         Universitat Politècnica de Catalunya, Barcelona \\
         {\tt\{casimiro.pio.carrino, marta.ruiz, jose.fonollosa\}@upc.edu}}

\abstract{
Recently, multilingual question answering became a crucial research topic, and it is receiving increased interest in the NLP community. However, the unavailability of large-scale datasets makes it challenging to train multilingual QA systems with performance comparable to the English ones. In this work, we develop the Translate Align Retrieve (TAR) method to automatically translate the Stanford Question Answering Dataset (SQuAD) v1.1 to Spanish. We then used this dataset to train Spanish QA systems by fine-tuning a Multilingual-BERT model. Finally, we evaluated our QA models with the recently proposed MLQA and XQuAD benchmarks for cross-lingual Extractive QA. Experimental results show that our models outperform the previous Multilingual-BERT baselines achieving the new state-of-the-art value of 68.1 F1 points on the Spanish MLQA corpus and 77.6 F1 and 61.8 Exact Match points on the Spanish XQuAD corpus. The resulting, synthetically generated SQuAD-es v1.1 corpora, with almost 100\% of data contained in the original English version, to the best of our knowledge, is the first large-scale QA training resource for Spanish.}
\begin{document}
\maketitleabstract

\section{Introduction}
Question answering is a crucial and challenging task for machine-reading comprehension and represents a classical probe to assesses the ability of a machine to understand natural language \cite{DBLP:journals/corr/HermannKGEKSB15}. In the last years, the field of QA has made enormous progress, primarily by fine-tuning deep pre-trained architectures \cite{NIPS2017_7181,DBLP:journals/corr/abs-1810-04805,DBLP:journals/corr/abs-1907-11692,DBLP:journals/corr/abs-1906-08237} on large-scale QA datasets. Unfortunately, large and high-quality annotated corpora are usually scarce for languages other than English, hindering advancement in Multilingual QA research. \\
Several approaches based on cross-lingual learning and synthetic corpora generation have been proposed. 
Cross-lingual learning refers to zero, and few-shot techniques applied to transfer the knowledge of a QA model trained on many source examples to a given target language with fewer training data.
\cite{Artetxe:etal:2019,DBLP:journals/corr/abs-1907-06042,liu-etal-2019-xqa}
On the other hand, synthetic corpora generation methods are machine-translation (MT) based designed to automatically generate language-specific QA datasets as training resources \cite{alberti-etal-2019-synthetic,lee-etal-2018-semi,DBLP:journals/corr/TureB16}. Additionally, a multilingual QA system based on MT at test time has also been explored \cite{DBLP:journals/corr/abs-1809-03275} \\\\
In this paper, we follow the synthetic corpora generation direction. In particular, we developed the Translate-Align-Retrieve (TAR) method, based on MT and unsupervised alignment algorithm to translate an English QA dataset to Spanish automatically. Indeed, we applied our method to the popular SQuAD v1.1 generating its first Spanish version. We then trained a Spanish QA systems by fine-tuning the Multilingual-BERT model. Finally, we evaluated our model on two Spanish QA evaluation set taken from the. Our improvements over the current Spanish QA baselines demonstrated the capability of the TAR method, assessing the quality of our synthetically generated translated dataset.\\\\
In summary, the contributions we make are the followings: i) We define an automatic method to translated the SQuAD v1.1 training dataset to Spanish that can be generalized to multiple languages. ii) We created SQuAD-es v1.1, the first large-scale Spanish QA. iii) We establish the current state-of-the-art for Spanish QA systems validating our approach. We make both the code and the SQuAD-es v1.1 dataset freely available\footnote{https://github.com/ccasimiro88/TranslateAlignRetrieve}.
\section{The Translate-Align-Retrieve (TAR) Method on SQuAD}
This section describes the TAR method and its application for the automatic translation of the Stanford Question Answering Dataset (SQuAD) v.1.1 \cite{DBLP:journals/corr/RajpurkarZLL16} into Spanish. The SQuAD v1.1 is a large-scale machine reading comprehension dataset containing more than 100,000 questions crowd-sourced on Wikipedia articles. It represents a high-quality dataset for extractive question answering tasks where the answer to each question is a span annotated from the context paragraph. It is partitioned into a training and development set with 80\% and 10\% of the total examples, respectively. Unlike the training set, the development set contains at least two answers for each posed question intended for robust evaluation. Each example in the SQuAD v.1.1 is a $(c, q, a)$ tuple made of a context paragraph, a question, and the related answers along with its start position in the context $a_{start}$.\\
Generally speaking, the TAR method is designed for the translation of the source $(c_{src}, q_{src}, a_{src})$ examples into the corresponding target language $(c_{tgt}, q_{tgt}, a_{tgt})$ examples. It consists of three main components:
\begin{itemize}
    \item A trained neural machine translation (NMT) model from source to target language
    \item A trained unsupervised word alignment model
    \item A procedure to translate the source $(c_{src}, q_{src}, a_{src})$ examples and retrieve the corresponding $(c_{tgt}, q_{tgt}, a_{tgt})$ translations using the previous components
\end{itemize}
The next sections explain in details how we built each component for the Spanish translation of the SQuAD v1.1 training set.

\subsection{NMT Training}
We built the first TAR component training from scratch an NMT model for English to Spanish direction. Our NMT parallel corpus is created by collecting the en-es parallel data from several resources. We first collected data from the WikiMatrix project \citelanguageresource{DBLP:journals/corr/abs-1907-05791} which uses state-of-the-art multilingual sentence embeddings techniques from the LASER toolkit\footnote{https://github.com/facebookresearch/LASER} \cite{DBLP:journals/corr/abs-1811-01136,DBLP:journals/corr/abs-1812-10464} to extract N-way parallel corpora from Wikipedia. Then, we gathered additional resources from the open-source OPUS corpus avoiding domain-specific corpora that might produce textual domain mismatch with the SQuAD v1.1 articles. Eventually, we selected data from 5 different resources, such as Wikipedia, TED-2013, News-Commentary, Tatoeba and OpenSubTitles \citelanguageresource{lison-tiedemann-2016-opensubtitles2016,DBLP:journals/corr/WolkM15c,TIEDEMANN12.463}.\\
The data pre-processing pipeline consisted of punctuation normalization, tokenisation, true-casing and eventually a joint source-target BPE segmentation \cite{sennrich-etal-2016-neural} with a maximum of 50k BPE symbols. Then, we filtered out sentences longer than 80 tokens and removed all source-target duplicates. The final corpora consist of almost 6.5M parallel sentences for the training set, 5k sentence for the validation and 1k for the test set. The pre-processing pipeline is performed with the scripts in the Moses repository\footnote{https://github.com/moses-smt/mosesdecoder} and the Subword-nmt repository\footnote{https://github.com/rsennrich/subword-nmt}.\\ We then trained the NMT system with the Transformer model \cite{NIPS2017_7181}. We used the implementation available in OpenNMT-py toolkit \cite{article} in its default configuration for 200k steps with one GeForce GTX TITAN X device. Additionally, we shared the source and target vocabularies and consequently, we also share the corresponding source an target embeddings between the encoder and decoder.
After the training, our best model is obtained by averaging across the final three consecutive checkpoints. \\Finally, we evaluated the NMT system with the BLEU score \cite{papineni-etal-2002-bleu} on our test set. The model achieved a BLEU score of 45.60 point showing that the it is good enough to be used as a pre-trained English-Spanish translator suitable for our purpose. 
\subsection{Source-Translation Context-Alignment}
The role of the alignment component is to compute the alignment between the context sentences and their translations. We relied on an efficient and accurate unsupervised word alignment called \textit{eflomal} \cite{Ostling2016efmaral} based on a Bayesian model with Markov Chain Monte Carlo inference. We used a fast implementation named \textit{efmaral} \footnote{https://github.com/robertostling/efmaral} released by the same author of eflomal model. The implementation also allows the generation of some priors that are used at inference time to align quickly. Therefore, we used tokenized sentences from the NMT training corpus to train a token-level alignment model and generate such priors.
\subsection{Translate and Retrieve the Answers}
The final component, defines a strategy to translate and retrieve the answers translation to obtain the translated version of the SQuAD Dataset. Giving the original SQUAD Dataset textual content, three steps are applied:
\begin{enumerate}
    \item Translate all the $(c_{src}, q_{src}, a_{src})$ instances 
    \item Compute the source-translation context alignment $align_{src-tran}$
    \item Retrieve the answer translations $a_{tran}$ a using the source-translation context alignment $align_{src-tran}$
\end{enumerate}
The next sections describes in details the steps below.\\\\
\textbf{Content Translation and Context-Alignment}
The first two steps are quite straightforward and easy to describe. First of all, all the $(c_{src}, q_{src}, a_{src})$ examples are collected and translated with the trained NMT model. Second, Each source context is split into sentences, and then the alignments between the context sentences and their translations are computed. Before the translation, each context source $c_{src}$ is split into sentences. Both the final context translation $c_{tran}$ and context alignment $align(c_{src},c_{tran})$ are consequently obtained by merging the sentence translations and sentence alignments, respectively. Furthermore, it is important to mention that, since the context alignment is computed at a token level, we computed an additional map from the token positions to the word positions in the raw context. The resulting alignment maps a word position in the source context to the corresponding word position in the context translation.  \\
Eventually, each source context $c_{src}$ and question $q_{src}$ is replaced with its translation $c_{tran}$ and $q_{tran}$ while the source answer $a_{src}$ is left unchanged. To obtain the correct answer translations, we designed a specific retrieval mechanism implemented in the last TAR component described next.
\\\\
\textbf{Retrieve Answers with the Alignment}
The SQUAD Dataset is designed for extractive question answering models where each question must be an answer that belongs to the paragraph. It poses a significant constraint to take into account when data are translated automatically. We found that in most cases, the translation of a paragraph is different from the translation of the answer that is contained in it. It may occur because a neural generative model, like the NMT, produces the translation of word conditioned on its context\\
To overcome this issue, we leverage on the previously computed context alignment $align_{src-tran}$. Therefore, we designed an answer extraction procedure that retrieves answers even when the answer translation is not contained in the paragraph translation. First, we use the $align_{src-tran}$ to map the word positions of the answer source $(a_{src}^{start},...,a_{src}^{end})$ to the corresponding word positions of the answer translation $(a_{tran}^{start},...,a_{src}^{end})$. Also, a position reordering is applied to extract the $a_{tran}^{'start}$ and the $a_{tran}^{'end}$ as the minimum and maximum over $(a_{tran}^{start},...,a_{src}^{end})$, respectively. This position reordering accounts for the inversion during the translation.
Then, for a given context, we look up the answer translation $a_{tran}$ as a span of the context translation $c_{tran}$. The span it is searched from the corresponding start position $a_{tran}^{start}$ in the context translation $c_{tran}$. It is necessary to detect in which part of the context translation $c_{tran}$ the answer translation $a_{tran}$ is mapped, to prevent the extraction of the incorrect answer span when it appears more than one sentence. Furthermore, the $a_{tran}$ is lower-cased to improve the matching probability on the $c_{tran}$. If the answer translated is found in context translation, it is retrieved. In the opposite case, we retrieve the answer translation $a_{tran}$ as the span between the $a_{tran}^{start}$ and $a_{tran}^{end}$. The pseudo-code in figure \ref{pseudo_code} shows the algorithm implementation.
\begin{algorithm}
\SetAlgoLined
\KwResult{$a_{tran}$}
\For {$c_{src}$ in contexts paragraph}{
  $c_{tran}$ $\leftarrow$ get  context translation \;
  $align(c_{src}, c_{tran})$ $\leftarrow$ get context alignment\;
   \For {$q$ in questions}{
   \For {$a_{src}$ in answers}{
     $a_{tran}$ $\leftarrow$ get answer translation \;
     \uIf{$a_{tran}$ in  $c_{tran}$}
     {return $a_{tran}$\;}
     \uElse{$(a_{src}^{start},...,a_{src}^{end})$ $\leftarrow$ get src positions\;
     compute tran positions with alignment\\
     $(a_{tran}^{start},...,a_{tran}^{end})$ $\leftarrow align_{src-tran}$\;
     compute start position as minimum\\
     $a_{tran}^{'start}\leftarrow min(a_{tran}^{start}, a_{tran}^{end})$\;
     compute end position as maximum \\
     $a_{tran}^{'end}\leftarrow max(a_{tran}^{start}, a_{tran}^{end})$\;
     return $a_{tran}\leftarrow c_{tran}[a_{tran}^{'start}:a_{tran}^{'end}$]\;
  }}}
}
\caption{Implementation of the answer retrieval with alignment for each $(c, q, a)$ example. The $c_{src}$ and $c_{tran}$ are the context source and translation, $q_{src}$ is the question source,  $a_{src}^{start}$, $a_{src}^{end}$ and  $a_{tran}^{start}$, $a_{tran}^{end}$  are the start and end positions for the answer source and the answer translation, $align_{src-tran}$ is the source-translation context alignment, and $a_{tran}$ is the retrieved answer.}
\label{pseudo_code}
\end{algorithm}

\section{The SQuAD-es Dataset}
We applied the TAR method to both the  SQuAD v1.1 training datasets. The resulting Spanish version is referred as SQuAD-es v.1.1. In table \ref{cqa} we show some $(c_{es}, q_{es}, a_{es})$ examples taken from the SQuAD-es v1. These examples show some good and bad $(c, q, a)$ translated examples giving us a first glimpse of the quality of the TAR method for the SQuAD dataset. 

\subsection{Error Analysis}
As follows, we conduct a detailed error analysis in order to better understand the quality of the translated $(c_{es}, q_{es}, a_{es})$ data.

The quality of the translated $(c_{es}, q_{es}, a_{es})$ examples in the SQuAD-es v1.1 Dataset depends on both the NMT model and the unsupervised alignment model. The interplay among them in the TAR method that determines the final result. Indeed, while bad NMT translations irremediably hurt the data quality also an erroneous source-translation alignment is responsible for the retrieval of wrong answer spans. 

Therefore, we carried out a qualitative error analysis on SQuAD-es v1.1 in order to detect and characterize the produced errors and the factors behind them.
We inferred the quality of the $(c_{es}, q_{es}, a_{es})$ examples by looking at errors in the answer translations $\{a_{es}\}$. Indeed, based on the TAR method, a wrong answer translation provides an easy diagnostic clue for a potential error in both the context translation $c_{es}$  or the source-translation context alignment $align_{en-es}$. We also make use of the question translations $\{q_{es}\}$ to asses the level of correctness of the answer translations $\{a_{es}\}$. We pointed out systematic errors and classified them in two types: 
\paragraph{Type I: Misaligned Span} The answer translation is a span extracted from a wrong part of the context translation and indeed does not represent the correct translation of the source answer. This error is caused to a misalignment in the source-translation alignment, or a translation error when the NMT model produces a context translation shorter than the source context that consequently generates a wrong span.
\paragraph{Type II: Overlapping Span} The answer translation is a span with a certain degree of overlap with the golden span on the context translation. Indeed, it might contain some additional words or punctuation, such as trailing commas or periods. In particular, the additional punctuation is present when the source annotation exclude punctuation while words in the span contain these character, but the source annotation does not. Sometimes, we also found that the answer translation span overlaps part of the next sentence. This error type is generated by a slightly imprecise source-translation alignment or by an NMT translation error. Nevertheless, we noticed that often enough the resulting answer translation, respect to its question, turns out to be acceptable. Table \ref{cqa_error_type} shows some examples of error type.\\

Overall, the two factors for these error types are both the NMT component and the alignment component in our TAR method. In order to have more quantitative idea of how the error types are distributed across the $(c_{es}, q_{es}, a_{es})$ examples, we randomly selected an article from SQuAD v1.1 and manually counted the error types. Besides, we divided the total examples into two sets, depending on how the answer translations are retrieved. The first set contains all the $(c_{es}, q_{es}, a_{es})$ instances, while the second, smaller set, contains only the $(c_{es}, q_{es}, a_{es})$ instances retrieved when the answer translation is matched as span in the context translation. In this way, we isolate the effect of the alignment component in the answer retrieval and evaluate its impact on the distribution of the error types. Table \ref{error_type_analysis}  shows the number of occurrence for each error type on the two sets of $(c_{es}, q_{es}, a_{es})$ examples.
\begin{table}[!h]
\begin{center}
\begin{tabularx}{\columnwidth}{l|X|X}
      \hline
     \textbf{Error } \# (\%) & \textbf{$\{(c, q, a)\}_{all}$}& \textbf{$\{(c, q, a)\}_{no\_align}$}\\
      \hline
      Type I&  15 (7\%)&0 (0\%)\\
      Type II&98 (43\%) &7 (10\%) \\
      Total&113 (50\%) & 7 (10\%)\\
            \hline
      Type II (acc.)&68 &4\\
      \# of $(c,q,a)$ ex. & 227&67\\
\end{tabularx}
\caption{\label{error_type_analysis}The number of error types occurrences and its percentage for two sets of $(c_{es}, q_{es}, a_{es})$ instances retrieved with and without alignment.}
 \end{center}
\end{table}

As a result, we found that the alignment is responsible for the introduction of a large number of error occurrences in the translated  $(c_{es}, q_{es}, a_{es})$ instances. As a consequence, when the answer translations are retrieved with the source-translation alignment, we found a significant increase of 40\% of the total errors. On the other side, when the alignment is not involved in the retrieval process, the total number of translated examples is drastically reduced by 70\% of the total number of examples in the other case. However, the number shows that the relative percentage of acceptable answers increased when the alignment is used. This analysis indicate that the TAR method can produce two kinds of a synthetical dataset, a bigger one with noisy examples and a smaller one with high quality. In the next section, we generate two Spanish translation of the SQuAD v1.1 training dataset, by considering or not $(c_{es}, q_{es}, a_{es})$ retrieved with the alignment, to empirically evaluate their impact on the QA system performance.
\subsection{Cleaning and Refinements}
After the translation, we applied some heuristics to clean and refine the retrieved $(c_{es}, q_{es}, a_{es})$ examples. Based on the error analysis, we post-processed the type II errors.  In particular, we first filter out words in the answers translations belonging to the next sentences. Then, we removed the extra leading and trailing punctuation. Eventually, we also removed empty answers translations that might be generated during the translation process. \\Moreover, in order to examine the impact on the QA system performance, we produced two versions of the SQuAD-es v1.1 training dataset. A standard one, containing all the translated $(c_{es}, q_{es}, a_{es})$ examples and referred to as SQuAD-es v1.1 and another that keep only the  $(c_{es}, q_{es}, a_{es})$  examples retrieved without the use of the alignment, named SQuAD-es-small.
Table \ref{SQuAD-es_statistics} shows the statistics of the final SQuAD-es v1.1 datasets in terms of how many $(c_{es}, q_{es}, a_{es})$ examples are translated over the total number of examples in its original English version. We also show the average context, question and answer length in terms of token. As a result, we the SQuAD-es v1.1 training contains almost the 100\%  of the SQuAD v1.1 data while the SQuAD-es-small v1.1 is approximately half the size, with a about 53\% of the data.  In the next section, these two SQUAD-es v1.1 datasets will be employed to train Spanish question Answering models.
\begin{table}[!h]
\begin{center}
\begin{tabularx}{\columnwidth}{l|X|X}
      \hline
      &  \textbf{SQuAD-es} &  \textbf{SQuAD-es-small}\\
      \hline
    \# of ex. &  87595/87599 & 46260/87599  \\
      Avg. $c$ len & 140 & 138\\
      Avg. $q$ len & 13& 12\\
      Avg. $a$ len & 4 &3\\
\end{tabularx}
\caption{\label{SQuAD-es_statistics}Number of $(c_{es}, q_{es}, a_{es})$ examples in the final SQuAD-es v1.1 Datasets over the total number of the original SQuAD v.1.1 $(c_{en}, q_{en}, q_{en})$. We also computed the average length for the context, question and answers in terms of tokens.}
 \end{center}
\end{table}

\begin{table*}[ht]
\begin{center}
\begin{tabular}{lp{6.8cm}p{6.8cm}}
      \hline
      &\textbf{en}&\textbf{es}\\
\hline
Context&Fryderyk Chopin was born in \textbf{Żelazowa Wola}, 46 kilometres (29 miles) west of Warsaw, in what was then the Duchy of Warsaw, a Polish state established by Napoleon. The parish baptismal record gives his birthday as \textbf{22 February 1810}, and cites his given names in the Latin form \textbf{Fridericus Franciscus} (in Polish, he was Fryderyk Franciszek). However, the composer and his family used the birthdate 1 March,[n 2] which is now generally accepted as the correct date.& Fryderyk Chopin nació en \textbf{Żelazowa Wola,} 46 kilómetros al oeste de Varsovia, en lo que entonces era el Ducado de Varsovia, un estado polaco establecido por Napoleón. El registro de bautismo de la parroquia da su cumpleaños el \textbf{22 de febrero de 1810,} y cita sus nombres en latín \textbf{Fridericus Franciscus} (en polaco, Fryderyk Franciszek). Sin embargo, el compositor y su familia utilizaron la fecha de nacimiento 1 de marzo, [n 2] que ahora se acepta generalmente como la fecha correcta.
\\\\
 Question&1) In what village was Frédéric born in?&1) ¿En qué pueblo nació Frédéric?\\
Answer&1) \textbf{Żelazowa Wola}&1) \textbf{Żelazowa Wola,} \\\\
Question&2) When was his birthday recorded as being?&2) ¿Cuándo se registró su cumpleaños?\\
Answers&2) \textbf{22 February 1810}&2) \textbf{22 de febrero de 1810,}\\\\
Question&3) What is the Latin form of Chopin's name?&3) ¿Cuál es la forma latina del nombre de Chopin?\\
Answer&3) \textbf{Fridericus Franciscus}&3)\textbf{ Fridericus Franciscus}
\\\\
\hline
Context&During the Five Dynasties and Ten Kingdoms period of China \textbf{(907–960}), while the fractured political realm of China saw no threat in a Tibet which was in just as much political disarray, there was little in the way of Sino-Tibetan relations. Few documents involving Sino-Tibetan contacts survive from the Song dynasty (960–1279). The Song were far more concerned with countering northern enemy states of \textbf{the Khitan}-ruled Liao dynasty (907–1125) and \textbf{Jurchen}-ruled Jin dynasty (1115–1234).&Durante el período de las Cinco Dinasties y los Diez Reinos de China \textbf{(907-960),} mientras que el reino político fracturado de China no vio ninguna amenaza en un Tíbet que estaba en el mismo desorden político. Pocos documentos que involucran contactos sino-tibetanos sobreviven de la dinastía Song (960-1279). Los Song estaban mucho más preocupados por contrarrestar los estados enemigos del norte de \textbf{la dinastía Liao gobernada por Kitán} (907-1125) y la dinastía Jin \textbf{gobernada por Jurchen} (1115-1234).\\\\
Question&1) When did the Five Dynasties and Ten Kingdoms period of China take place?& 1) ¿Cuándo tuvo lugar el período de las Cinco Dinasties y los Diez Reinos de China?\\
Answer&1) \textbf{907–960}&1) \textbf{(907-960),}\\\\
Question&2) Who ruled the Liao dynasty?&2) ¿Quién gobernó la dinastía Liao?\\
Answer&2) \textbf{the Khitan}&2) \textbf{la dinastía Liao gobernada por Kitán}\\\\
Question&3) Who ruled the  Jin dynasty?&3) ¿Quién gobernó la dinastía Jin?
\\
Answer&3) \textbf{Jurchen}&3) \textbf{gobernada por Jurchen }
\\
\hline
\end{tabular}
\caption{\label{cqa}Examples of $(c_{es}, q_{es}, a_{es})$ examples selected in two of the first articles in the  SQuAD v1.1 training dataset, \textit{Frédéric Chopin} and \textit{Sino-Tibetan relations during the Ming dynasty}. The corresponding answer spans in the contexts are highlighted in bold.}
\end{center}
\end{table*}

\begin{table*}[!h]
\begin{center}
\begin{tabular}{lp{6.6cm}p{6.6cm}l}
      \hline
      &\textbf{en}&\textbf{es}&Error\\
\hline
Context&The Medill School of Journalism has produced notable journalists and political activists including \textbf{38} Pulitzer Prize laureates. National correspondents, reporters and columnists such as The New York Times's Elisabeth Bumiller, David Barstow, Dean Murphy, and Vincent Laforet, USA Today's Gary Levin, Susan Page and Christine Brennan, NBC correspondent Kelly O'Donnell, CBS correspondent \textbf{Richard Threlkeld}, CNN correspondent Nicole Lapin and former CNN and current Al Jazeera America anchor Joie Chen, and ESPN personalities Rachel Nichols, Michael Wilbon, Mike Greenberg, Steve Weissman, J. A. Adande, and Kevin Blackistone. The bestselling author of the A Song of Ice and Fire series, George R. R. Martin, earned a B.S. and M.S. from Medill. Elisabeth Leamy is the recipient of 13 Emmy awards  and 4 Edward R. Murrow Awards.&La Escuela Medill de Periodismo ha producido notables periodistas y activistas políticos incluyendo \textcolor{red}{38 premios Pulitzer. Los} corresponsales nacionales, reporteros y columnistas como Elisabeth Bumiller de The New York Times, David Barstow, Dean Murphy, y Vincent Laforet, Gary Levin de USA Today, Susan Page y Christine Brennan de \textcolor{red}{NBC.} A. Adande, y Kevin Blackistone. El autor más vendido de la serie Canción de Hielo y Fuego, George R. R. Martin, obtuvo un B.S. Y M.S. De Medill. Elisabeth Leamy ha recibido 13 premios Emmy y 4 premios Edward R. Murrow.&-\\\\
 Question&1) Which CBS correspondant graduated from The Medill School of Journalism? &1) ¿Qué corresponsal de CBS se graduó de la Escuela Medill de Periodismo?&-\\
Answer&1) \textbf{Richard Threlkeld}&1) \textcolor{red}{NBC.} &Type I\\\\
Question&2) How many Pulitzer Prize laureates attended the Medill School of Journalism?&1) ¿Cuántos premios Pulitzer asistieron a la Escuela Medill de Periodismo?&-\\
Answers&2) \textbf{38}&2) \textcolor{red}{38 premios Pulitzer. Los}& Type II\\\\
\hline
 Context&Admissions are characterized as "\textbf{most selective}" by U.S. News \& World Report. There were 35,099 applications for the undergraduate class of 2020 (entering 2016), and 3,751 (\textbf{10.7\%}) were admitted, making Northwestern one of the most selective schools in the United States. For freshmen enrolling in the class of 2019, the interquartile range (middle 50\%) on the SAT was 690–760 for critical reading and 710-800 for math, ACT composite scores for the middle 50\% ranged from 31–34, and \textbf{91\%} ranked in the top ten percent of their high school class.& Las admisiones se caracterizan como \textcolor{green}{"más selectivas"} por U.S. News \& World Report. Hubo 35.099 solicitudes para la clase de pregrado de 2020 (ingresando en 2016), y 3.751 \textcolor{green}{(10.7\%)} fueron admitidos, haciendo de Northwestern una de las escuelas más selectivas en los Estados Unidos. Para los estudiantes de primer año de matriculación en la clase de 2019, el rango intermedio (50\% medio) en el SAT fue de 690-760 para lectura crítica y 710-800 para matemáticas, compuesto ACT para el \textcolor{red}{31\%}.&-\\\\

 Question&1) What percentage of freshman students enrolling in the class of 2019 ranked in the top 10\% of their high school class?& 1) ¿Qué porcentaje de estudiantes de primer año matriculados en la clase de 2019 se ubicó en el 10\% superior de su clase de secundaria?&-\\
 Answer& 1) \textbf{91\%}& 1) \textcolor{red}{31\%}&Type I\\\\
Question &2) What percentage of applications were admitted for the undergraduate class entering in 2016?&2)    ¿Qué porcentaje de solicitudes fueron admitidas para la clase de pregrado en 2016?
&-\\
Answer&2)\textbf{ 10.7\%}&2) \textcolor{green}{(10.7\%)}&Type II (acc.)\\\\
Question&3) How selective are admissions at Northwestern characterized by U.S. News & 3) ¿Cuán selectivas son las admisiones en Northwestern caracterizadas por U.S. News&-\\
Answer&3) \textbf{most selective}&3) \textcolor{green}{"más selectivas"}&Type II (acc.)\\
\hline
\end{tabular}
\caption{\label{cqa_error_type}Examples of error types in the $(c_{es}, q_{es}, a_{es})$ examples in the randomly selected article \textit{ Northwestern University} from SQuAD v1.1. Dataset. Wrong answers are highlighted in red while acceptable answers in green.}
\end{center}
\end{table*}

\section{QA Experiments}
We trained two Spanish QA models by fine-tuning a pre-trained Multilingual-BERT (mBERT) model on our SQuAD-es v1.1 datasets following the method used in \cite{DBLP:journals/corr/abs-1810-04805}. We employed the implementation in the open-source HuggingFace's Transformers library \cite{Wolf2019HuggingFacesTS}. Our models have been trained for three epochs one GTX TITAN X GPU device using the default parameter's values set in the HugginFace scripts.

The goal is to assess the quality of our synthetically generated datasets used as a training resource for Spanish QA models. We performed the Spanish QA evaluation on two recently proposed, freely available, corpus for cross-lingual QA evaluation, the MLQA and XQuAD corpora \cite{lewis2019mlqa,Artetxe:etal:2019}. The MLQA corpus is an N-way parallel dataset consisting of thousands of QA examples crowd-sourced from Wikipedia in 7 languages, among which Spanish. It is split into development and test sets, and it represents an evaluation benchmark for cross-lingual QA systems. 

Similarly, the XQuAD corpus is another cross-lingual QA benchmark consisting of question-answers pairs from the development set of SQuAD v1.1 translated by professional translators in 10 languages, including Spanish. Therefore, we used both the MLQA and XQuAD benchmark to evaluate the performance of our trained models. We measured the QA performance with the F1, and Exact Match (EM) score computed using the official evaluation script available in the MLQA repository, that represents a multilingual generalization of the commonly-used SQuAD evaluation script \footnote{https://rajpurkar.github.io/SQuAD-explorer/}.
\begin{table*}[ht]
\begin{center}
\begin{tabular}{l|l|l}

      \hline
      &\textbf{F1 / EM} &\textbf{es} \\
       \hline \hline
      Our&TAR-train + mBERT (SQuAD-es)& \textbf{68.1 / 48.3} \\
      models&TAR-train + mBERT (SQuAD-es-small)& 65.5 / 47.2 \\
      \hline
      MLQA&mBERT & 64.3 / 46.6 \\
      mBERT baselines&Translate-train + mBERT & 53.9 / 37.4\\
      \hline
      state-of-the-art &XLM (MLM + TLM, 15 languages) &68.0 / 49.8\\
      \hline
\end{tabular}
\caption{\label{MLQA_eval}Evaluation results  in terms of F1 and EM scores on the MLQA corpus for our Multilingual-BERT models trained with two versions of SQuAD-es v1.1 and the current Spanish Multilingual-BERT baselines }
\end{center}
\end{table*}

\begin{table*}[ht]
\begin{center}
\begin{tabular}{l|l|l}
      \hline
      &\textbf{F1 / EM} & \textbf{es} \\
       \hline\hline
     Our &TAR-train + mBERT (SQuAD-es) & \textbf{77.6} /\textbf{ 61.8}\\
     models &TAR-train + mBERT (SQuAD-es-small) & 73.8 / 59.5\\
      \hline
      &JointMulti 32k voc   & 59.5 / 41.3\\
      XQuAD&JointMulti 200k voc  (state-of-the-art)& 74.3 / 55.3 \\
      mBERT baselines&JointPair with Joint voc & 68.3 / 47.8 \\
      &JointPair with Disjoint voc  &72.5 / 52.5 \\
      \hline
\end{tabular}
\caption{\label{XQuAD_eval}Evaluation results in terms of F1 and EM scores on the XQuAD corpus for our Multilingual-BERT models trained with two versions of SQuAD-es v1.1 and the current Spanish Multilingual-BERT baselines }
\end{center}
\end{table*}
Results of the evaluation are shown in Table \ref{MLQA_eval} and Table \ref{XQuAD_eval}. On the MLQA corpus, our model best model beat the state-of-the-art F1 score of the XLM (MLM + TLM, 15 languages)\cite{DBLP:journals/corr/abs-1901-07291} baseline for Spanish QA. Equally, on the XQuAD corpus, we set the state-of-the-art for F1 and EM score. Overall, the TAR-train+mBERT models perform better than the current mBERT-based baselines, showing significant improvements of $1.2\%-3.3\%$ increase in F1 score and $0.6\%-6.5\%$ raise in EM points across the MLQA and XQuAD benchmarks. Interestingly enough, when compared to a model trained on synthetical data, such as the translate-train-mBERT, our TAR-train-mBERT models show even more substantial improvements, setting out a notable increase of  $11.6-14.2\%$ F1 score and  $9.8-10.9\%$ EM score. These results indicate that the quality of the SQuAD-es v1.1 dataset is good enough to train a Spanish QA model able to reach state-of-the-art accuracy, therefore proving the efficacy of the TAR method for synthetical corpora generation.\\
The QA evaluation demonstrates that the performance on the Spanish MLQA and XQuAD benchmarks of the mBERT increased by $2.6-4.2\%$ F1 score and $1.1-2.3\%$ EM score when the SQuAD-es v1.1 dataset is used compared the SQuAD-es-small v1.1 dataset. Based on the error analysis in section 3, we can assume that the SQuAD-es v1.1 is a bigger but noisy dataset, compared to the SQuAD-es-small that is the smaller but more accurate. Therefore, we observe that the mBERT model may be robust enough to tolerate noisy data giving more importance to the quantity. This observation connects to the problem of quality versus quantity in synthetical corpora generation and its application to multilingual reading comprehension \cite{lee-etal-2019-learning}

\section{Conclusions}
In this work we have designed a TAR method designed to automatically translate the SQuAD-es v1.1 training dataset to Spanish. Hence, we applied the TAR method to generated the SQuAD-es v1.1 dataset, the first large-scale training resources for Spanish QA. Finally, we employed the SQuAD-es v1.1 dataset to train QA systems that achieved state-of-the-art perfomance on the Spanish QA task, demonstrating the efficacy of the TAR approach for synthetic corpora generation. Therefore, we make the SQuAD-es dataset freely available and encourage its usage for multilingual QA.

The results achieved so far encourage us to look forward and extend our approach in future works. First of all, we will apply the TAR method to translated the SQuAD v2.0 dataset \cite{DBLP:journals/corr/abs-1806-03822} and other large-scale extractive QA such as Natural Questions\cite{47761}. Moreover, we will also exploit the modularity of the TAR method to support languages other than Spanish to prove the validity of our approach for synthetic corpora generation.

\section*{Acknowledgements}

This work is supported in part by the Spanish Ministerio de Econom\'ia y Competitividad, the European Regional  Development  Fund  and  the  Agencia  Estatal  de  Investigaci\'on,  through  the  postdoctoral  senior grant Ram\'on y Cajal, the contract TEC2015-69266-P (MINECO/FEDER,EU) and the contract PCIN-2017-079 (AEI/MINECO).

\section{Bibliographical References}\label{reference}

\bibliographystyle{lrec}
\bibliography{paper.bib}

\section{Language Resource References}
\label{lr:ref}
\bibliographystylelanguageresource{lrec}
\bibliographylanguageresource{paper.bib}
\end{document}